# SEAM PUCKERING OBJECTIVE EVALUATION METHOD FOR SEWING PROCESS

BRAD Raluca[1], HĂLOIU Eugen[2], BRAD Remus[2]

[1] Lucian Blaga University of Sibiu, Romania, Department of Industrial Machinery and Equipment, Faculty of Engineering, B-dul Victoriei 10, 550024 Sibiu, Romania, E-Mail: raluca.brad@ulbsibiu.ro

[2] Lucian Blaga University of Sibiu, Romania, Department of Computer Science and Electrical Engineering, Faculty of Engineering, B-dul Victoriei 10, 550024 Sibiu, Romania, E-Mail: eugen.haloiu@ulbsibiu.ro, remus.brad@ulbsibiu.ro

Corresponding author: Brad, Raluca, E-mail: raluca.brad@ulbsibiu.ro

*Abstract: The paper presents an automated method for the assessment and classification of puckering defects detected during the preproduction control stage of the sewing machine or product inspection. In this respect, we have presented the possible causes and remedies of the wrinkle nonconformities. Subjective factors related to the control environment and operators during the seams evaluation can be reduced using an automated system whose operation is based on image processing. Our implementation involves spectral image analysis using Fourier transform and an unsupervised neural network, the Kohonen Map, employed to classify material specimens, the input images, into five discrete degrees of quality, from grade 5 (best) to grade 1 (the worst).*

## 1. QUALITY AND AUTOMATION IN THE TEXTILE INDUSTRY

The textile industry is one of the traditional and dynamic sectors where the customer quality requirements are constantly changing as a result of trends in fashion and the development of production tools. In order to satisfy clients' demands, the variables that affect product quality must be kept under control during the production cycle: design, manufacturing, delivery and maintenance.

The evaluation process of a sewn product relating to appearance and performance have to rely on a holistic perspective that includes both fabrics and sewing threads assessment, but also consider their interactions during sewing, wearing and maintenance of the product. Throughout the manufacturing process, the woven, non-woven or knitted fabrics are controlled from two to more than six times in order to detect the defects which may occur, followed by their classification and if possible, remedying. A good compatibility between sewing thread and materials will influence the product quality and productivity. Otherwise, during the sewing process, the fabric is damaged or the machine stops at unanticipated time intervals [1].

After execution, a correct seam need to be smooth and flat, without puckering, tuck developing or seam damage, having an appropriate behavior during pressing and cleaning. There are several standard test methods for evaluating the interactions between the threads and fabrics after the execution of the seams and stitching. Some procedures assess seam strength, slippage, failure, damage, pucker and jamming before and after cleaning [2]. ISO 7770, AATCC 88 B and AATCC 143 [3,4,5] standards use sets of images and rating scales in order to evaluate the appearance of seams, using grades from 1 (worse) to 5 (best quality seam). Inspectors should compare the stitching samples with the standard images, in different environments, which cause subjective results (figure 1).

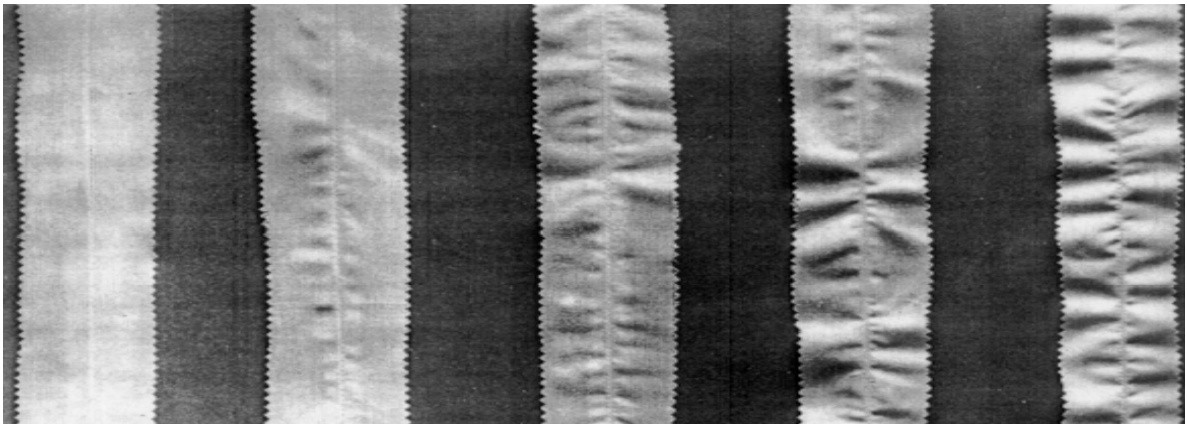
*Fig. 1: Standard images for the subjective assessment of seam pucker [4].*

Due to long reaction time and fatigue of the human operator, an automatic inspection would be able to verify and classify the seams with a much higher speed and would eliminate the subjective factor. The system can be used both in the pre-production stage, for sewing machines adjustment, and also in product inspection. The ability to recognize flaws and stop production immediately after the occurrence of the defect is important for clothing manufacturers.

The automatic control system may use different technologies for image acquisition, containing mechanical components, computer software, video cameras, lighting and video equipment. In particular, an automatic defect detection is be based on an electro-optical device for fabric surface inspection using a two-dimensional scanner of the warp and weft directions, or on a complex of video camera and uniform illumination source connected to video acquisition system [6].

## 2. THE SEAM PUCKERING DEFECT: CAUSES AND REMEDIES

The seam puckering phenomenon is defined as a local defect of a clothing item in the form of large ridges of material beside the seam and is considered one of the most serious defects in garment manufacturing [7]. The complete elimination of this flaw during pressing operations is almost impossible, and therefore in practice, it is often to accept a lower grade as normal. Consequently, the objective assessment of seam puckering is essential, such that the final product will be acceptable to the client [8].

Seam pucker may also be described as a differential shrinkage that arises throughout the seam line and is caused by the instability of the seam. Usually, wrinkling appears due to improper selection of stitching parameters and material properties, leading to an inequality of fabrics lengths that are sewn together and affecting the appearance characteristics. In serious cases, puckering can appear as a wave coming from the seams. Since sewing operation is subjected to excessive tensions, it produces a stretching of sewing threads, leading to an extension over the whole surfaces of the fabrics.

Although sewing threads have usually a controlled elasticity, they are overstretched when large tensions are implied in process. After sewing, the threads tend to relax, trying to return to the original length. As the stitches shrink, wrinkles appear in the material and can not be detected immediately, but in a later stage. The threads used in sewn products must also have a good stability to washing and ironing, as differential shrinkage between sewing thread and fabrics may cause puckering.

Other fabrics characteristics which affect seam stability and puckering are fabric density and structure. By stitching, threads snap the fiber material into a new position, inserting into material structure and tending to change it. This phenomenon is more obvious when the fabric is made of fine, dense and low resistance to compression yarns. In the case of differences in fiber composition, fabrics structure, extensibility and stability, puckers may occur due to feeding failure. In order to avoid this situation, it is necessary to adjust the presser foot pressure to a minimum value. The feeding systems used in stitching control are: a positive or negative differential conveyor, consisting of two teeth elements in front and behind the needle with adjustable amplitude, and a simple oriented tooth feed.

In order to reduce puckering, sewing machine and stitching parameters are adjusted. In the case of the sewing machine, adjustments are made on the conveyor mechanism, tension control and needle selection. Regarding the processing parameters, the stitching step should be as small as possible, while the value of cutting and sewing angle must be correlated with yarn and fabric structure. Using a similar fabric fiber composition thread with thermo stability, low elongation and recovery, puckers can be decreasead or avoided.

## 3. AN OBJECTIVE SEAM PUCKERING EVALUATION METHOD

Subjective evaluation methods have the disadvantage a higher assessment time, differences between appraisals, partiality towards certain colors or models, and training needs. In the attempt of objectification, two SP synthetic indicators may be used for seam pucker description: related to width and to the length. [9] The hypothesis assumes that following wrinkled defect occurrence, the thickness of two layers sewn assembly increases, while the length decreases, comparing to the initial length of unraveled material. In these conditions, the following formula can be used:

$$SP = \frac{t_s - 2t}{2t} * 100 \ (\%) \quad \text{or} \quad SP = \frac{l - l_s}{l_s} * 100 \ (\%) \tag{1}$$

where $t_s$ = seam thickness, $t$ = fabric thickness, $l$ = length of unraveled fabric, $l_s$ = length of sewn assembly.

However, incoherence and time-consuming are noticed for these assessments.

Image processing techniques have been already applied in the textile industry. Research was carried out to investigate the cross section of fibers, yarn structure, yarn thickness [10], texture fault detection, seam pucker etc. Image processing is sometimes combined with a classification/recognition step achieved using neural networks. A large review paper on this field, including many textile applications, has been published by [11].

The assessment of seams is one of the research topics in the aim of textile industry automation. The foundations of this field start with a geometric modeling of puckering and a review of the methods and techniques available at that time, in the view of measurement [12]. In [13], the use of a k nearest neighbor classifier achieved an 81% rate of successful classification rate compared with human experts. This implementation is trying to improve a previous one from the same authors, presented in [14]. A wavelet based detectors of surface smoothness or wrinkles and puckering defects have been used by [15], joined with a 3D scanning system.

Our implementation was derived from [16] and involves a spectral image analysis using Fourier transform, and an unsupervised neural network, Kohonen Map to classify material specimens, which are the input images into five discrete degrees of quality, to grade 5 (best) to grade 1 (the worst). A similar approach was proposed using fractal theory [17]. The learning and testing stages are depicted in figure 2. The next paragraphs will present the basics of our processing scheme.

### 3.1. The Otsu Algorithm for image binarization

In image processing, the Otsu's method is used to perform image binarization, in order to separate objects from background. The algorithm assumes that the image to be segmented contains two classes of pixels (bimodal histogram) and calculates the optimum threshold to separate the classes. The frequencies of grey levels are established and the probabilities for each possible threshold level are computed. The variance of the pixels levels on either side of the threshold will be estimated, both for the object and the background region. Among all possible values, the threshold that minimizes the inter-class variance will be selected, being defined as a weighted sum of the two classes' variances [18]:

$$s_w^2(t) = w_1(t)s_1^2(t) + w_2(t)s_2^2(t) \tag{2}$$

where $w_{1,2}$ represents the probabilities of the two classes separated by threshold $t$, and $s_{1,2}^2$ symbolizes the variance of the two classes. The average of each class is given by the weighted average of frequency intensities, with $L$ the number of grey levels:

$$m_1(t) = \sum_{i=1}^{t} \frac{ip(i)}{w_1(t)} \quad \text{and} \quad m_2(t) = \sum_{i=t+1}^{L} \frac{ip(i)}{w_2(t)} \tag{3}$$

The individual variances of the classes are:

$$s_1^2(t) = \sum_{i=1}^{t} [i - m_1(t)]^2 \frac{p(i)}{w_1(t)} \quad \text{and} \quad s_2^2(t) = \sum_{i=t+1}^{L} [i - m_2(t)]^2 \frac{p(i)}{w_2(t)} \tag{4}$$

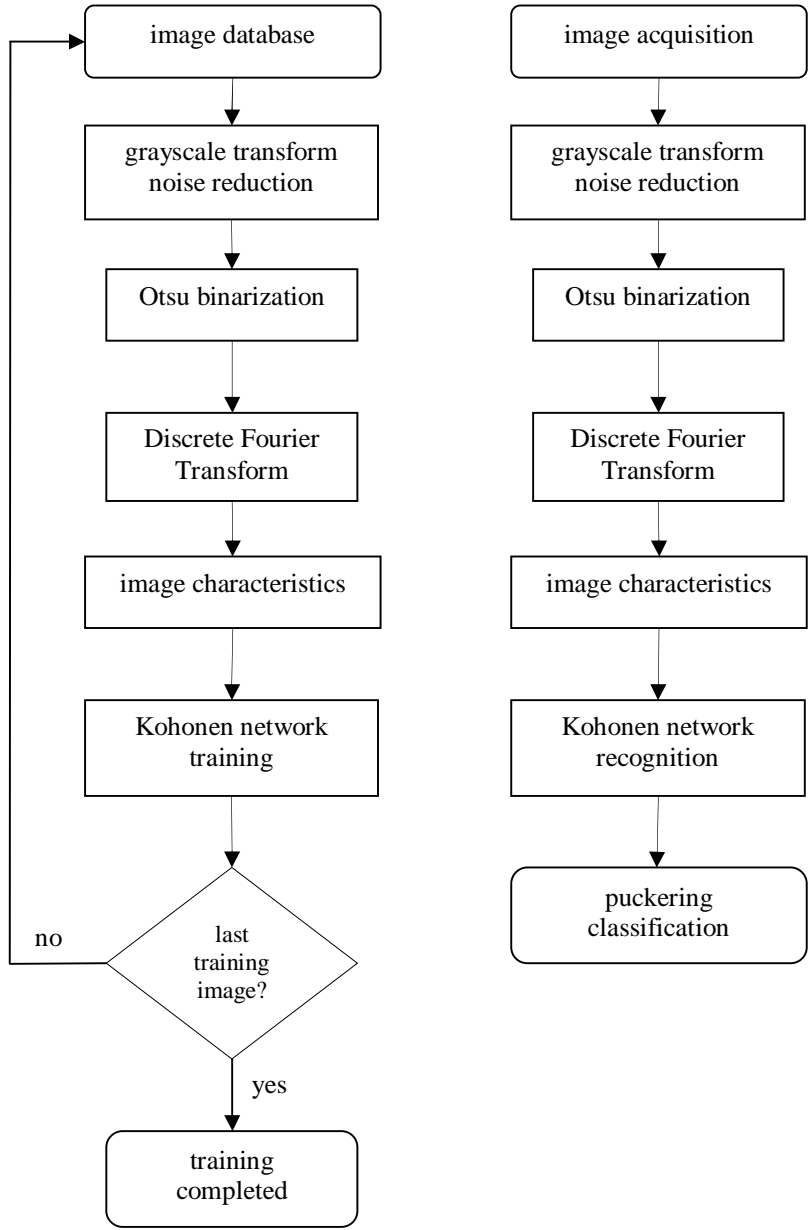

*Fig. 2: The proposed image processing framework.*

### 3.2. The Fourier transform

The Fourier Transform is an important tool in image processing, representing the input spatial domain image in the frequency field and therefore, each point in the output representing a particular frequency contained in the image. It is used in a wide range of applications, such as image analysis, image filtering, image reconstruction and image compression. In the context of the present paper, the Discrete Fourier Transform (DFT) will be used. The DFT is actually a sampled Fourier transform and will not contain all frequencies being present in the image, but only a set of samples large enough to describe the spatial domain image, as the number of frequencies corresponds to the number of pixels [19].

For an $N \times N$ image, the DFT is given by:

$$F(k,l) = \frac{1}{N^2} \sum_{m=0}^{N-1} \sum_{n=0}^{N-1} f(m,n) e^{-i2p\left(\frac{km}{N}+\frac{ln}{N}\right)} \qquad (5)$$

where $f(m, n)$ is the spatial domain image. Each pixel value is multiplied with an exponential term, called the basis function, and summed over the domain. $F(0, 0)$ represents the continuous component of the image (ie the mean gray level), while $F(N-1, N-1)$ characterizes the larger frequency. In the same way, the DFT image can be retransformed in the spatial domain as an inverse transform:

$$f(m,n) = \frac{1}{N^2} \sum_{k=0}^{N-1} \sum_{l=0}^{N-1} F(k.l)e^{-i2p\left(\frac{km}{N}+\frac{ln}{N}\right)} \quad (6)$$

where $\frac{1}{N^2}$ is a normalization term.

The double sum in equation 1 must be computed for each point (*k*, *l*). However, due to the separabilty of the Fourier transform, the above formula can be expressed in two N series of one-dimensional transforms, decreasing the number of computations:

$$F(k,l) = \frac{1}{N} \sum_{n=0}^{N-1} P(k.n)e^{-i2p\frac{ln}{N}} \quad (7)$$

where

$$P(k,n) = \frac{1}{N} \sum_{m=0}^{N-1} f(m,n)e^{-i2p\frac{km}{N}} \quad (8)$$

The DFT produces an output which can be displayed using two images: either the real and imaginary part or the magnitude and phase. In image processing, the amplitude values are usually employed since information about the geometrical structure of the image space is carried out. Nevertheless, in order to compute the inverse DFT after processing the frequency domain, both amplitude and phase array of values must be preserved and stored on wide representation variables [19].

### 3.3. The Kohonen Map

The Kohonen networks are also known as "self-organizing maps", a special type of artificial neural networks trained using unsupervised learning in order to produce a discrete representation of the input space. Self-organizing maps are different from other neural networks as they use a function to preserve neighborhoods topological properties of the input space. Therefore, Kohonen maps are used to approximate the distribution of input vectors, the dimensionality reduction while maintaining data in the vicinity or for clustering.

Kohonen maps are organized in two layers. The first level of the network is the input layer, while the second is the competitive level, organized as an array. The two layers are fully interconnected; each input node is connected to all nodes in the competitive layer, as in figure 3 [20].

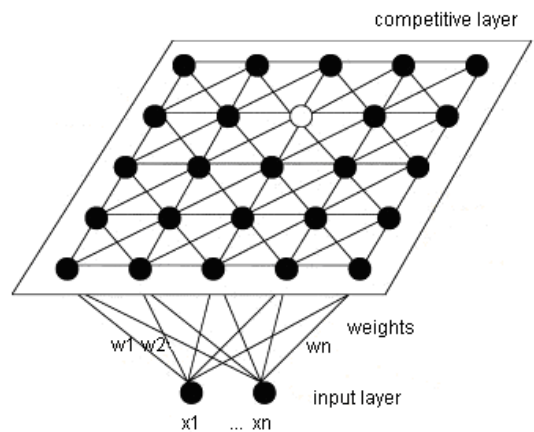

*Fig. 3: The basic structure of the Kohonen network [20].*

Each connection has an associated weight. In the initial state, the network weights have random values in the range [0, 1]. The input pattern is an *n* dimensional vector. As a result, the input models are uniformly distributed over a square. The first step in the operation of Kohonen maps is to compute a matching value for each node in the competitive layer. This value measures the extent to which the weight of each node corresponds to the input node. The matching value is the distance between *X* (inputs) and *W* (weights) vectors:

$$\|X - W_i\| = \sqrt{\sum_j (x_j - w_{ij})^2} \quad (9)$$

The node with the best match wins the competition. This node is set as follows:

$$\|X - W_c\| = \min_i \{\|X - W_i\|\} \quad (10)$$

where *c* is the best node.

After the winning node is identified, the next step is to identify its neighborhood, as those nodes in a square centered on the winning node. The weights are updated for all neurons that are in the neighborhood of the winning node. The update equation is:

$$\Delta u_{ij} = \begin{cases} a(e_j - u_{ij}), & \text{if node } i \text{ is in the neigborhood} \\ 0 & \text{else} \end{cases} \quad (11)$$

where $\alpha$ is the learning step, initially ($\alpha_0$) in the range [0.2, 0.5]. During the learning process, the value is decreased over iterations until 0, using:

$$a_i = a_0 \left(1 - \frac{t}{T}\right) \quad (12)$$

where *t* is the current iteration and *T* the total number of iterations.

Besides the learning step, the neighborhood size must also be updated over the iterations in the learning process, as in equation 13.

$$\begin{aligned} c - d < x < c + d \\ c - d < y < c + d \end{aligned} \quad (13)$$

with *c* representing the winning node, *d* the distance from *c* to the edge of the neighborhood, and *x, y* the node coordinates. Usually, the initial value of *d* is chosen as half to three quarters the size of the competitive layer and updated using:

$$d_i = d_0 \left(1 - \frac{t}{T}\right) \quad (14)$$

where *t* is the current iteration and *T*, the total number of iterations.

In summary, the basic rules of the Kohonen network can be described as follows [20]:
- § Locate the unit in the competitive layer whose weight fits best with the input,
- § Update the weights of the selected unit and its neighbors, thus increasing the level of matching,
- § Gradually decreases the neighborhood size and adjust the weights during the iterations of the learning process

## 4. RESULTS

The input image is represented by a sewn sample image with puckering. At the beginning of the learning or testing framework, a preprocessing block was introduced, due to his main role in filtering. As in seam defects detection, color is not necessary; the acquired images are transformed to grayscale (figure 4).

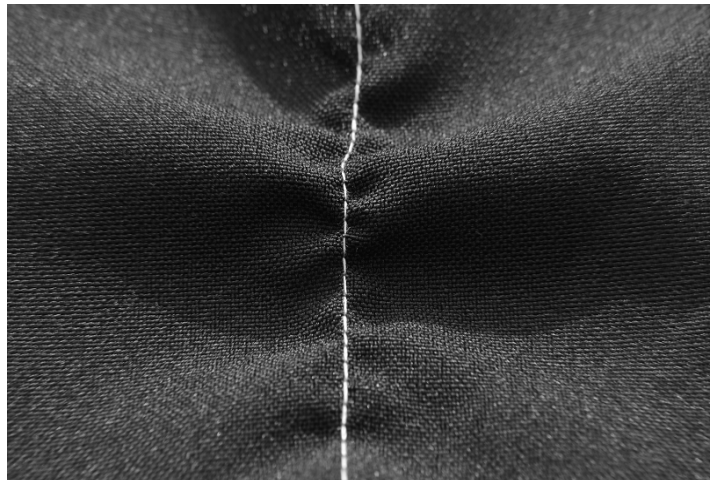

***Fig. 4:** Grayscale image of a sewn presenting puckering.*

The result of a DFT is represented by the amplitude and phase of frequency components of the input image. The amplitude shows how much of a particular frequency is being presented in the image, while the phase represents how the signal is offset from the origin, or particularly, how much the sinus wave is shifted to the left or the right, as shown in figure 5.

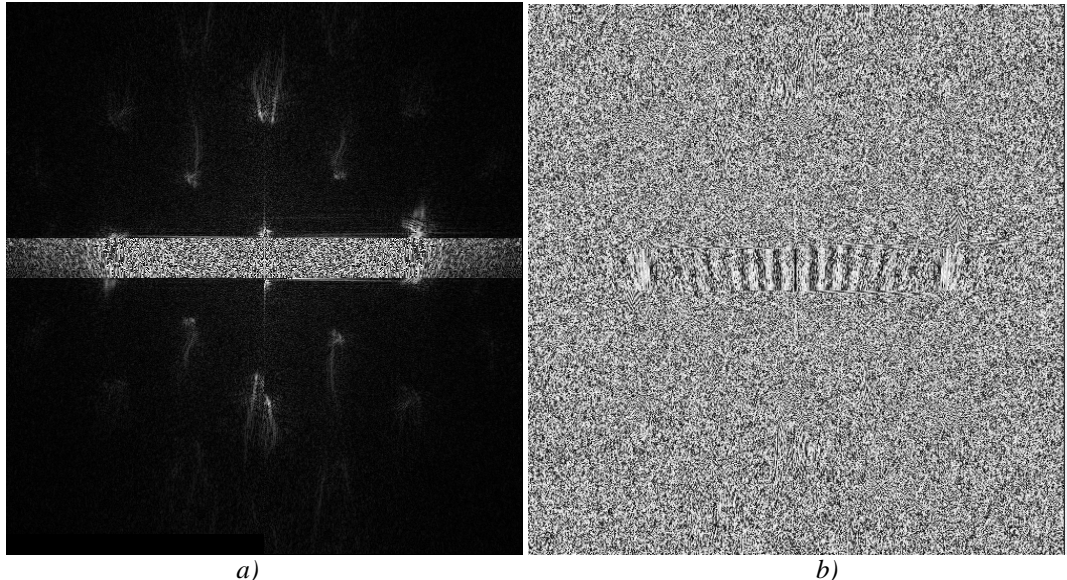

*a)* *b)*
***Fig. 5:*** *The amplitude a) and phase b) result of the above image DFT.*

Our classification approach of puckering is based on series of samples that have been categorized accordance to some human experts. In this sense, the puckering features present in the learning and test images have been pre-classified using the seam puckering quality standard. In the case of the proposed framework, the image features are extracted using spectral analysis by Fourier transform and the results stored in an array. We have down-sampled the corresponding array, in order to fit into the low resolution Kohonen Map, of a 100x100 size. The network training stage will consist in presenting five input vectors (derived from the down-sampled arrays), representing the puckering grades. The network learning stops when the error falls below a value that is very close to zero. The puckering classification consists in providing an input vector derived from the image supposed to be classified. A scalar product between the input values vectors and the weighted training images is computed. The result will be assigned to one of the five classes of which the input image belongs.

This method requires the image acquisition of samples in special conditions. An oblique light is applied on the material and, due to the wrinkled material appearance, the shadows will highlight the nonconformities of the fabric surface.

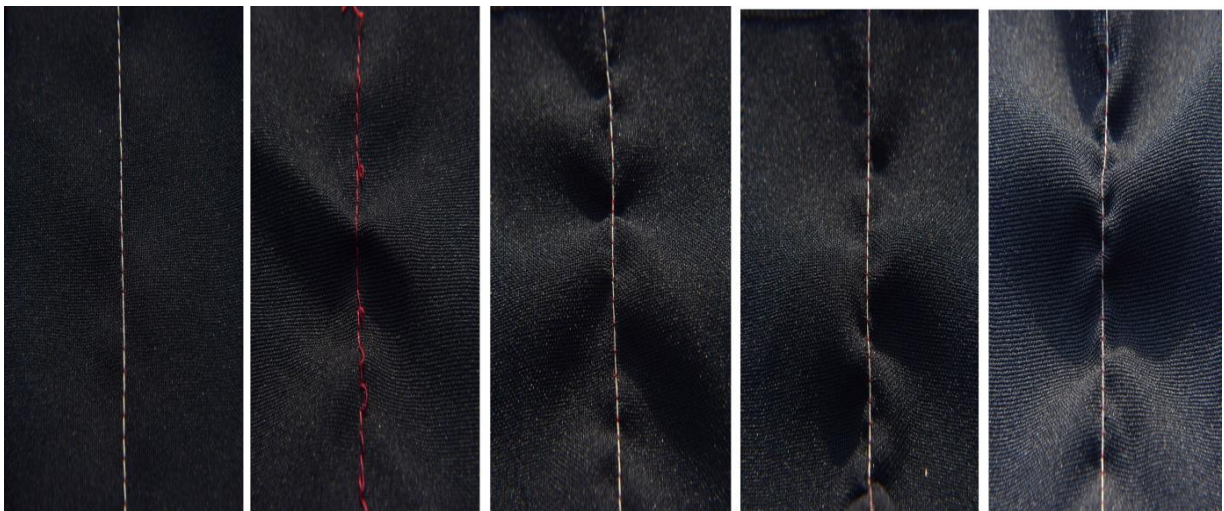

***Fig. 6:*** *The training set.*

The investigation was performed on 26 samples of 100% polyester plain woven fabric, with 21 cm$^{-1}$ warp and weft density, and 170 grams/sqm specific weight. Specimens were cut into 10x10 cm sizes and midst sewn in pairs using 301 type lockstitch and a 4 cm$^{-1}$ seam density. Two colors of sewing thread were used: white to the needle and red to the bobbin, with 80 Nm sewing thread count, 100% PES.

The network was trained using the features images from figure 6, extracted using the DFT. In order to test the functionality of the network another set of 21 test images was used (figure 7). These images were originally classified subjectively using visual information. The classification results using the Kohonen network are as follows:
- 71.42% of the images were classified correctly.
- 28.58% of the images were misclassified.

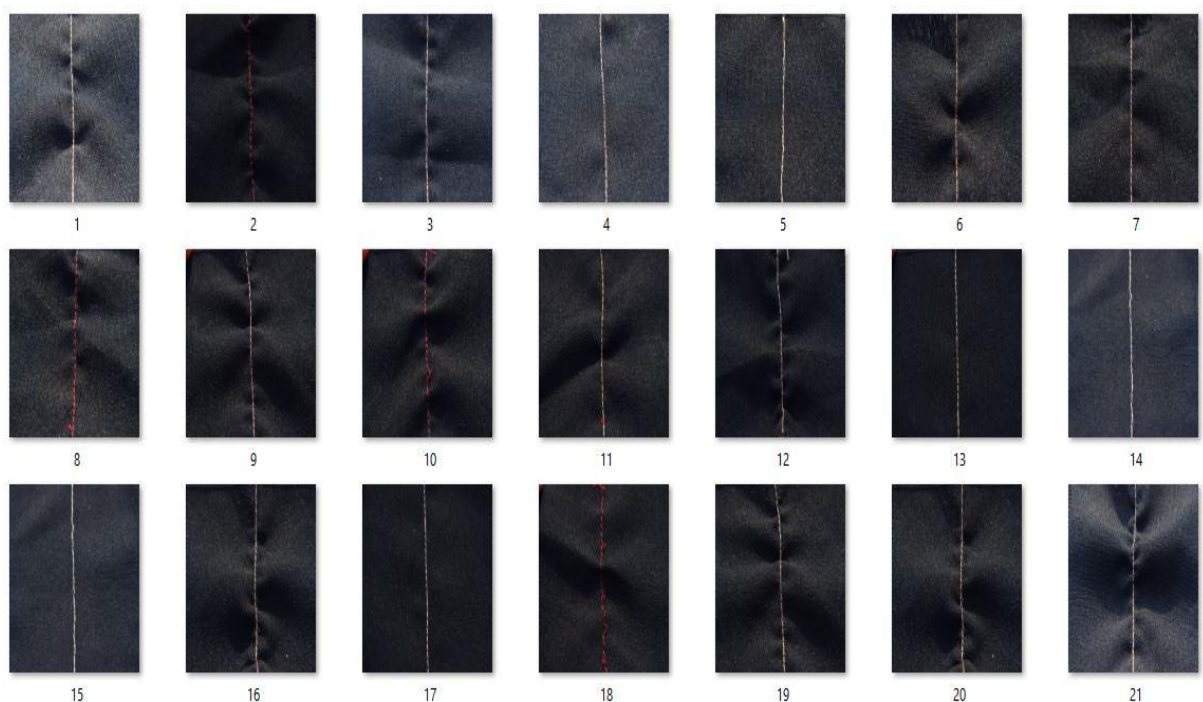

*Fig. 7: The testing set.*

## 5. CONCLUSIONS

The applicability domain of the presented framework is in the textile industry, namely seams quality control and sewn assemblies classification in terms of visual quality. Various algorithms have been applied on the acquired images in order to improve their processability. Currently, defect detection is done using input images containing only horizontal seams. For further development of the application, fault detection will be completed on several types of stitches and seam shapes, for example, circular.

The image classification is done using visual information, based on subjective standard images. We have created five quality classes using puckering images, used in the training of the neural network. By processing a large number of samples, classification could be redefined and improved.

The detection of seams defects will be investigated using three images categories, containing horizontal seams from clothing manufacturing, airbag assembly and automotive upholstery.